\documentclass[a4paper,twocolumn,10pt]{article}

\usepackage{epsfig}
\usepackage{subcaption}
\usepackage{calc}
\usepackage{amssymb}
\usepackage{amstext}
\usepackage{amsmath}
\usepackage{amsthm}
\usepackage{multicol}
\usepackage{pslatex}
\usepackage{apalike}
\usepackage{enumitem}
\usepackage[bottom]{footmisc}
\usepackage{algorithm2e}
\usepackage{booktabs}
\usepackage{url}
\usepackage{tikz}
\usetikzlibrary{calc, shapes, backgrounds, positioning, arrows.meta, fit}

\tikzset{
  blk/.style   = {draw, rounded corners=2pt, minimum height=8mm,
                  minimum width=13mm, align=center, font=\scriptsize, fill=gray!8},
  stage/.style = {blk, fill=blue!8, minimum width=11mm},
  conv/.style  = {blk, fill=purple!8},
  head/.style  = {blk, fill=green!10, minimum width=17mm},
  chg/.style   = {blk, fill=orange!25, very thick, draw=orange!70!black},
  io/.style    = {font=\scriptsize\itshape, align=center},
  fl/.style    = {-{Latex[length=2mm]}, thick, gray!60!black},
  note/.style  = {font=\scriptsize\itshape, align=left, text=black!60},
}

\usepackage[a4paper,margin=2cm,columnsep=0.8cm]{geometry}
\usepackage[font=small,skip=4pt]{caption}
\begin{document}

\title{Supervision, Not Architecture: A Controlled Study of Transformer
Backbones for Local Feature Learning}

\author{Amit Sharma\\[2pt]
{\small Hamilton Institute, Maynooth University, Maynooth, Ireland}\\
{\small \texttt{amit.sharma.2021@mumail.ie}}\\[6pt]
{\small Accepted at VISAPP 2027, Valletta, Malta, February 2027 (\texttt{visapp.scitevents.org})}}
\date{}

\twocolumn[
\begin{@twocolumnfalse}
\maketitle
\begin{abstract}
Learned local feature detectors and descriptors remain
predominantly convolutional, even as transformers have displaced
convolutional backbones across dense prediction more broadly. The
strongest published transformer result for local features couples a
deformable-attention architecture with a change of training
supervision, leaving the two contributions confounded. We separate
them. Within a single pipeline built as a drop-in replacement for the
R2D2 backbone, we vary the attention mechanism with the supervision
held fixed, then vary the supervision with the architecture held
fixed, measuring throughout against a convolutional baseline that we
retrain ourselves on the same data. Our findings are
principally negative. Substituting a transformer backbone redistributes
matching accuracy rather than improving it, trading viewpoint accuracy
for illumination accuracy in a manner consistent with spatially
smoothed descriptors. Multi-scale deformable attention, the natural
architectural remedy for that smoothing, is worse than its
spatially-reduced counterpart on every measure under synthetic-warp
supervision and no better on any under depth-warped
supervision, and evidence from the training
dynamics argue against an implementation or training
failure. Replacing synthetic-warp supervision with depth-warped
correspondence on an unchanged architecture instead gains about $4.8$
AUC@$5^{\circ}$ on MegaDepth-1500 and closes about $40\%$ of the gap to
the convolutional baseline. We do not overtake that baseline on any
benchmark. We conclude that in this design space the supervision
pipeline, rather than the attention mechanism, is where performance
moves.
\end{abstract}
\bigskip
\end{@twocolumnfalse}
]

\section{\uppercase{Introduction}}
\label{sec:introduction}

\noindent Local feature detection and description underpins structure
from motion, visual localisation and SLAM. The learned methods that now
dominate the field, from SuperPoint~\cite{detone_superpoint_2018} to
R2D2~\cite{revaud_r2d2_2019}, are convolutional. As transformers have
displaced convolutional backbones across dense prediction, recent work
has begun to adopt them for local features as well, and the strongest
such result, RDD~\cite{chen_rdd_2025}, reports state-of-the-art
performance using a deformable transformer.

It is tempting to read that as evidence that the architectural
substitution is what helps. The evidence does not support the reading.
RDD departs from the R2D2 lineage along two axes simultaneously: it
adopts deformable attention, \emph{and} it trains on
MegaDepth~\cite{li_megadepth_2018} pairs under correspondence
supervision rather than on synthetically warped collections under
R2D2's objective. Its gains are equally consistent with an
architectural cause, a supervision cause, or any mixture of the two,
and to our knowledge the two have not been separated.

This paper separates them. We fix a single pipeline built as a drop-in
replacement for the R2D2 backbone, so that R2D2's losses, sampling and
extraction apply unchanged, and then vary one axis at a time: first the
attention mechanism with the supervision held constant, then the
supervision with the architecture held constant. Every configuration is
measured against a convolutional baseline we retrain ourselves on the
same data and a comparable iteration budget, so that no comparison
depends on a published number obtained under different conditions.

Our findings are principally negative, and we state them up front.
Replacing the convolutional backbone with a transformer does not
improve matching so much as redistribute it, trading viewpoint accuracy
for illumination accuracy in a manner consistent with spatially
smoothed descriptors. Multi-scale deformable attention, the natural
architectural remedy for that smoothing and the mechanism RDD adopts,
is worse than its spatially-reduced counterpart on all nine paired measurements we were able to make, across two supervision
pipelines. Changing the supervision instead, on an architecture held
fixed, gains about $4.8$~AUC@$5^{\circ}$ on MegaDepth-1500 and closes
about $40\%$ of the gap to the convolutional baseline. \textbf{We do not
overtake that baseline on any benchmark}, and we report this as an
analysis of where performance in this design space actually moves,
not as a competitive method.

We make three contributions. First, a controlled separation of the
architecture and supervision axes in local feature learning, on three
standard benchmarks against a self-retrained baseline. Second, an
architecture-level negative result for deformable attention that is
robust across two supervision pipelines, supported by evidence from
training dynamics that argues against an implementation or
training failure. Third, a supervision-level positive result showing
that depth-warped correspondence, applied to an unchanged architecture,
moves performance further than any architectural
substitution we tested.

\section{\uppercase{Related Work}}
\label{sec:related}

\paragraph{Learned local features.}
SuperPoint~\cite{detone_superpoint_2018} established the modern template
for jointly learned detection and description: a shared convolutional
encoder with separate detection and description heads, trained through
homographic adaptation on synthetically warped images.
R2D2~\cite{revaud_r2d2_2019} refined it by separating \emph{repeatability}
from \emph{reliability}, on the argument that a point being detected
consistently and a point being described distinctively are different
properties that a single confidence score conflates. R2D2 supplies the
loss family, the head structure and the training mix we build on, and
its convolutional backbone is the baseline our transformer replaces.
DISK~\cite{tyszkiewicz_disk_2020} pursues the same objective through
reinforcement learning over matching outcomes. All of these are
convolutional.

\paragraph{Hierarchical vision transformers.}
ViT~\cite{dosovitskiy_image_2021} applied a standard transformer encoder
to image patches, but produces a single-scale representation that suits
dense prediction poorly. Hierarchical designs followed.
PVT~\cite{wang_pvt_2021} introduced spatial-reduction attention, pooling
keys and values to a lower resolution so that attention remains tractable
at high spatial resolution; Swin~\cite{liu_swin_2021} instead restricts
attention to local windows that shift between layers.
SegFormer~\cite{xie_segformer_2021} combined spatial-reduction attention
with a mixed depthwise FFN to give the MiT backbone family, of which
MiT-B2 is the smallest variant that clearly outperforms ResNet-50 on
dense prediction. MiT-B2 is the backbone we adopt.

\paragraph{Deformable attention.}
Deformable attention~\cite{zhu2020deformable} replaces content-agnostic
spatial reduction with per-query learnable sampling: each query predicts
a small set of offsets and weights, and the value tensor is sampled at
those learned positions by bilinear interpolation. The defining property
is that \emph{where} aggregation happens is learned and content-dependent,
which is the inductive-bias opposite of spatial reduction's fixed
$R \times R$ pooling. That contrast is precisely why deformable attention
is the natural candidate for the smoothing problem we identify in
Section~\ref{sec:results:tradeoff}.

\paragraph{Transformers for matching and description.}
LoFTR~\cite{sun_loftr_2021} matches detector-free through interleaved
self- and cross-attention over dense CNN features;
RoMa~\cite{edstedt_roma_2024} pairs a frozen foundation-model encoder
with a transformer matching decoder. Closest to our setting,
RDD~\cite{chen_rdd_2025} adopts a deformable transformer for local
feature description and reports state-of-the-art results on HPatches and
MegaDepth-1500.

\paragraph{The gap this paper addresses.}
RDD's result is the strongest published evidence that deformable
attention benefits local feature learning, but it does not isolate the
mechanism. RDD differs from the R2D2 lineage along \emph{two} axes at
once: it adopts deformable attention, \emph{and} it trains on
MegaDepth~\cite{li_megadepth_2018} pairs under correspondence-supervised
losses rather than on synthetically warped collections under R2D2's
objective. Its reported gains are therefore consistent with an
architectural explanation, a supervision explanation, or any mixture of
the two, and the published literature does not, to our knowledge,
separate them.

This paper varies the two axes independently within a single fixed
pipeline. We hold the supervision constant and substitute the attention
mechanism, then hold the architecture constant and substitute the
supervision, and measure both against a convolutional baseline we
retrain ourselves on the same data. The result is that the two
axes are not of comparable weight, and that the one the literature has
attended to less is the one that moves performance.

\begin{figure*}[t]
\centering
\resizebox{\textwidth}{!}{%
\begin{tikzpicture}[node distance=4mm]
  \node[io] (in) {Image\\$H\times W$};
  \node[blk, right=of in]     (stem) {Stem\\$2\times$conv $3{\times}3$\\stride 4};
  \node[stage, right=of stem] (s1)   {Stage 1\\$\times 3$};
  \node[stage, right=of s1]   (s2)   {Stage 2\\$\times 4$};
  \node[stage, right=of s2]   (s3)   {Stage 3\\$\times 6$};
  \node[stage, right=of s3]   (s4)   {Stage 4\\$\times 3$};
  \node[blk, right=of s4]     (fpn)  {Multi-scale\\fusion};
  \node[blk, right=of fpn]    (up)   {Learned\\upsample\\$\times 4$};
  \node[head, above right=3mm and 6mm of up] (desc) {Descriptor\\$L_2$-norm, 128-d};
  \node[head, right=6mm of up]               (rel)  {Reliability\\$1{\times}1$ conv};
  \node[head, below right=3mm and 6mm of up] (rep)  {Repeatability\\$1{\times}1$ conv};
  \foreach \a/\b in {in/stem, stem/s1, s1/s2, s2/s3, s3/s4, s4/fpn, fpn/up}
    \draw[fl] (\a) -- (\b);
  \foreach \h in {desc, rel, rep} \draw[fl] (up.east) -- (\h.west);
  \begin{scope}[on background layer]
    \node[fit=(s1)(s4), draw=blue!40, dashed, rounded corners, inner sep=2mm,
      label={[font=\scriptsize\itshape,blue!50!black]above:MiT-B2 encoder}] {};
  \end{scope}
\end{tikzpicture}}
\caption{The shared pipeline with SR-attention in the encoder slot. A
two-layer convolutional stem reduces the input to stride $4$; the four
MiT-B2 encoder stages produce a feature pyramid; multi-scale fusion and
two learned upsampling blocks return the representation to full
resolution, where R2D2's three output heads are applied unchanged. Only
the encoder stages are substituted in the architectural comparison of
Section~\ref{sec:method:attention}.}
\label{fig:vis:base}
\end{figure*}
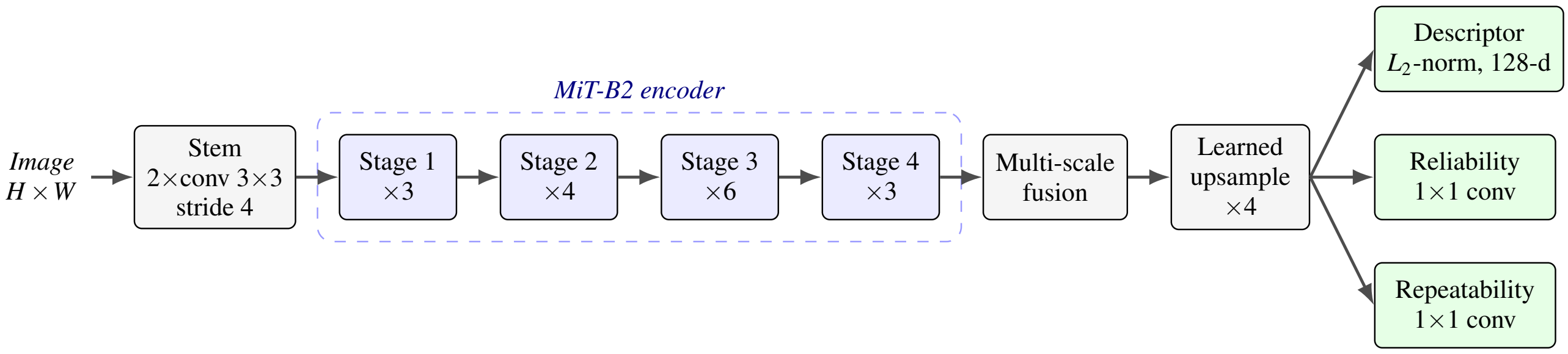

\begin{figure*}[t]
\centering
\resizebox{\textwidth}{!}{%
\begin{tikzpicture}[node distance=4mm]
  \node[io] (in) {Image};
  \node[blk, right=of in]     (stem){Stem\\stride 4};
  \node[stage, right=of stem] (s1) {Stage 1};
  \node[stage, right=of s1]   (s2) {Stage 2};
  \node[stage, right=of s2]   (s3) {Stage 3};
  \node[stage, right=of s3]   (s4) {Stage 4};
  \node[chg, below=9mm of $(s1)!0.5!(s4)$, minimum width=56mm] (de)
       {$1{\times}1$ projections to $d_{\mathrm{model}}=128$\\
        $+$ multi-scale deformable encoder ($4$ layers)};
  \node[blk, right=8mm of de.east] (fpn) {Multi-scale fusion\\($d_{\mathrm{model}}$)};
  \node[blk, right=of fpn] (up) {Learned\\upsample};
  \node[head, right=6mm of up] (h) {Descriptor\\Reliability\\Repeatability};
  \foreach \a/\b in {in/stem,stem/s1,s1/s2,s2/s3,s3/s4} \draw[fl] (\a) -- (\b);
  \foreach \n in {s1,s2,s3,s4} \draw[fl] (\n.south) -- (\n.south |- de.north);
  \draw[fl] (de) -- (fpn); \draw[fl] (fpn) -- (up); \draw[fl] (up) -- (h);
  \begin{scope}[on background layer]
    \node[fit=(s1)(s4), draw=blue!40, dashed, rounded corners, inner sep=1.5mm,
      label={[font=\scriptsize\itshape,blue!50!black]above:MiT-B2, pretrained}]{};
  \end{scope}
\end{tikzpicture}}
\caption{Variant~2: deformable refinement on top of MiT-B2. The
backbone is unchanged from Figure~\ref{fig:vis:base} and retains its
pretrained weights; the multi-scale deformable encoder is inserted
between the four stages and the multi-scale fusion, which is widened to
accept $d_{\mathrm{model}}$ channels at every level. Deformable
attention therefore \emph{refines} the SR-attention features here. In
Variant~1, by contrast, the deformable encoder replaces the transformer
stages outright, leaving the patch embeddings as fixed-stride feature
extractors.}
\label{fig:vis:v2}
\end{figure*}
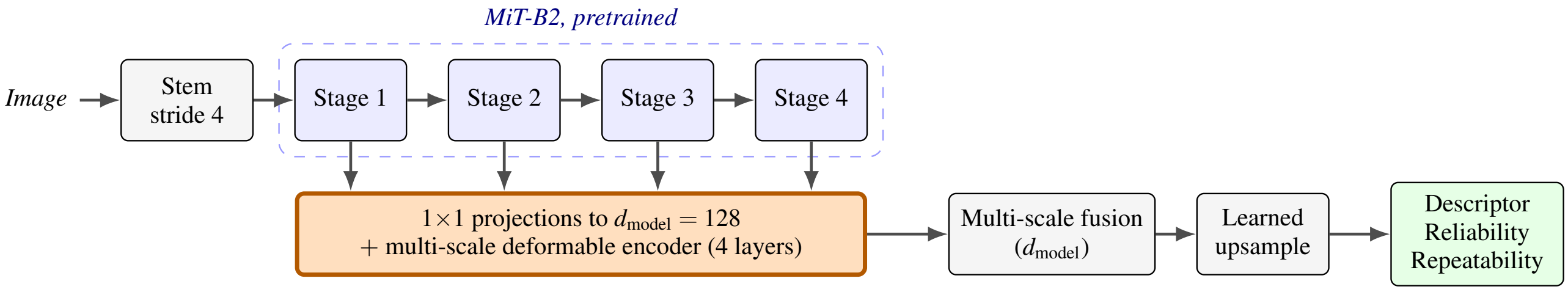

\section{\uppercase{Method}}
\label{sec:method}

\noindent Every configuration in this study shares one outer pipeline.
We describe it first, then the two attention mechanisms that are
substituted inside it, then the two supervision pipelines used to train
it. Holding the pipeline fixed is what allows the attention mechanism
and the supervision signal to be varied one at a time.

\subsection{Shared pipeline}
\label{sec:method:pipeline}

The network, shown in Figure~\ref{fig:vis:base}, is built as a drop-in
replacement for the R2D2 backbone within R2D2's own training framework.
It exposes the same descriptor, reliability and repeatability outputs as
the convolutional baseline, so R2D2's losses, sampling scheme and
extraction pipeline apply without modification and the backbone remains
the only changed variable.

A convolutional \textbf{stem} maps the input image to a feature map at
one quarter of the input resolution. We replace SegFormer's single
$7\times7$ stride-4 convolution with two $3\times3$ stride-2
convolutions, which reaches the same stride through smaller receptive
fields. The motivation is that corner-ness and edge orientation are the
signals a keypoint detector depends on, and a wide filter risks blurring
them at the first layer. The cost is that the stem cannot load
pretrained weights and is trained from scratch.

A four-stage \textbf{encoder} processes the stem output into feature
maps at $H/4$, $H/8$, $H/16$ and $H/32$. This is the component we
substitute, and Section~\ref{sec:method:attention} describes the two
mechanisms we place here.

A \textbf{multi-scale fusion} module projects all four stage outputs to
a common channel count with $1\times1$ convolutions, bilinearly resizes
them to the finest resolution, concatenates them and fuses them into a
128-channel map with a further $1\times1$ convolution. This is
SegFormer's all-MLP decoder with the per-pixel classification layer
replaced by the R2D2 output heads.

An \textbf{upsampling head} returns the fused map to input resolution
through two blocks, each a $\times 2$ bilinear upsample followed by a
$3\times3$ convolution with batch normalisation and ReLU. We avoid
transposed convolutions deliberately: their checkerboard
artefacts~\cite{odena_deconvolution_2016} would interfere with
sub-pixel keypoint localisation, and resize-then-convolve is the
remedy those authors recommend.

Three $1\times1$ \textbf{heads} produce the outputs. A two-way softmax
yields the reliability map, a second convolution yields the
repeatability map, and $L_2$ normalisation of the feature map yields the
descriptors. These are unchanged from R2D2.

\subsection{The two attention mechanisms}
\label{sec:method:attention}

\paragraph{SR-attention.} Our reference encoder is the MiT-B2
configuration of SegFormer~\cite{xie_segformer_2021}: four stages with
embedding dimensions $[64,128,320,512]$ and depths $[3,4,6,3]$, using
spatially-reduced attention, in which keys and values are pooled before
the attention product to make global attention tractable at high
resolution. The transformer blocks are initialised from the pretrained
MiT-B2 checkpoint. This is the configuration referred to throughout as
SR-attention.

\paragraph{Deformable attention.} Multi-scale deformable
attention~\cite{zhu2020deformable} replaces the pooled global product
with a small set of learned, content-dependent sampling offsets per
query, which is why it is a candidate remedy for the smoothing
described in Section~\ref{sec:results:tradeoff}: it samples sparsely and
adaptively rather than pooling uniformly. We use four encoder layers,
eight attention heads, eight sampling points per head per level and
$d_{\mathrm{model}} = 128$, following the Deformable DETR reference
configuration. We evaluate two ways of introducing it.

\emph{Variant 1} replaces the per-stage transformer blocks outright.
The four patch embeddings remain as fixed-stride feature extractors and
the deformable encoder becomes the only content-dependent transformation
in the backbone, giving 3.49\,M trainable parameters. We report this
variant for completeness but do not rest any conclusion on it, and the
reason should be stated plainly: it trains entirely from random
initialisation. The deformable encoder has no pretrained counterpart,
which is unavoidable, but the retained patch embeddings are
shape-identical to their MiT-B2 counterparts and are nonetheless left
random, because the loading routine is keyed to the per-stage block
structure this variant removes. Variant 1 is therefore confounded by
initialisation as well as by capacity.

\emph{Variant 2} (Figure~\ref{fig:vis:v2}) keeps the MiT-B2 backbone
intact, pretrained weights included, and inserts the deformable encoder
after the four stages and before the multi-scale fusion, which is
widened to accept $d_{\mathrm{model}}$ channels at every level. Deformable attention here
refines the SR-attention features rather than replacing them, and the
comparison is against an otherwise identical network without the
inserted encoder. \textbf{The architectural comparison in this paper
rests on Variant 2}, which is free of both confounds.

\subsection{The two supervision pipelines}
\label{sec:method:supervision}

Both pipelines supply the same thing to the same losses: a dense
correspondence field between an image pair. They differ only in where
that field comes from, which is what makes them comparable.

\paragraph{Synthetic warps (WASF).} The R2D2 training mix pairs web
images, Aachen day-time database images, day-to-night style-transfer
pairs and optical-flow pairs. Correspondence for the first three comes
from synthetic homographies applied to a single image, so the geometry
relating a pair is generated rather than observed.

\paragraph{Depth-warped correspondence.} We replace this with MegaDepth
image pairs in the covisibility overlap range $[0.3, 0.7]$, and build
the correspondence field from measured geometry: per-pixel depth,
relative pose and intrinsics warp each pixel into its partner view, and
a forward-backward consistency check masks pixels that are occluded or
disoccluded between the views. The pairs are real photographs of the
same scene from genuinely different viewpoints, so the supervision
contains the viewpoint variation that synthetic homographies cannot
represent.

The single controlled change between the two is the geometric source of
the correspondence field. Architecture, losses, batch size and crop size
are identical; the training length is not, since the depth-warped runs
receive about $1.5$ times as many optimiser iterations, as
Section~\ref{sec:setup:data} sets out.

\subsection{Losses}
\label{sec:method:losses}

We use R2D2's multi-loss family with one addition: the reliability average-precision
loss, the cosine-similarity loss and the peakiness loss, plus a hard-negative
triplet term at weight $0.5$ with margin $0.2$. Applying the same losses
across every transformer configuration is what isolates the backbone and the
supervision pipeline from any contribution of the objective. The
convolutional baseline does not carry the triplet term
(Section~\ref{sec:setup:baseline}).

A $3\times3$ sub-pixel offset head is present in the network but carries
zero weight in the loss and is used only as a diagnostic. It plays no
part in any result reported here.

\section{\uppercase{Experimental Setup}}
\label{sec:setup}

\subsection{Training}
\label{sec:setup:training}

The transformer blocks are initialised from the SegFormer MiT-B2
checkpoint in the mmsegmentation model zoo~\cite{mmseg_2020}, whose
encoder was pretrained on ImageNet-1k and fine-tuned as part of a full
SegFormer model on ADE20K~\cite{zhou_scene_2017}. The stem, the
multi-scale fusion, the upsampling head and the output heads are
initialised from scratch, the stem necessarily so because the
pretrained $7\times7$ patch embedding does not match our two-layer
$3\times3$ design.

Pretrained and scratch parameters receive different learning rates. The
pretrained blocks use a peak rate of $2\times10^{-5}$ to limit
catastrophic forgetting; scratch components use $2\times10^{-4}$. Both
groups use AdamW with weight decay $0.01$, a linear warm-up over the
first $5\%$ of steps, and cosine annealing thereafter.

The loss is R2D2's MultiLoss at unit weight on the reliability,
cosine-similarity and peakiness terms, with a hard-negative triplet term
at weight $0.5$ and margin $0.2$. The sub-pixel head carries weight
zero. This recipe is held fixed across every transformer configuration reported here
except where an intervention in Table~\ref{tab:vis:interventions}
explicitly varies it.

\subsection{Data and budget}
\label{sec:setup:data}

WASF-supervised runs use R2D2's training mix of web images, Aachen
day-time database images, day-to-night style-transfer pairs and
optical-flow pairs, at crop $192\times192$ and batch $16$ for $50$
epochs, which is approximately $50{,}000$ optimiser iterations.

Depth-supervised runs use MegaDepth~\cite{li_megadepth_2018} pairs in
the covisibility overlap range $[0.3, 0.7]$, at the same crop and batch
size for $3$ epochs. The pairs come only from the scenes in the training
list of the MegaDepth split distributed with LoFTR~\cite{sun_loftr_2021},
at most $2{,}700$ per scene and overlap bin, $405{,}491$ pairs in all;
the validation and test scenes, including the two from which
MegaDepth-1500 is built, are excluded. Because MegaDepth supplies far
more pairs per epoch, this is approximately $76{,}000$ iterations, about
$1.5$ times the WASF budget. An epoch-matched comparison would have
differed by more than an order of magnitude in gradient steps; the
remaining difference in iteration count is a confound on the
supervision comparison, which Section~\ref{sec:results:supervision}
notes where it applies.

We write pt$k$ for the checkpoint saved after epoch $k$, so the final
checkpoint of a $3$-epoch run is pt$2$. We use no validation set, so we
do not select checkpoints: every reported number is the final checkpoint
of its run.

\subsection{Baseline}
\label{sec:setup:baseline}

Comparisons are made against R2D2 retrained by us on the same WASF data, rather than against the published checkpoint. The
retrained model scores $0.304$ Overall MMA at threshold $1$ against the
published checkpoint's $0.321$, and matches it to within a point at
looser thresholds, so it reproduces the published baseline closely
enough to stand in for it. Using it removes the objection that any
transformer-versus-convolution gap might reflect a difference in
training data or schedule rather than in architecture.

The retrained baseline is not, however, matched to the transformer
configurations in every respect. It was trained with the reference R2D2
implementation, whose objective is the three-term MultiLoss without the
hard-negative triplet term; the transformer configurations carry that
term at weight $0.5$. Any transformer-versus-convolution comparison
therefore spans one additional loss term as well as the backbone. That
difference most likely favours the transformer: removing the triplet
term from \textsc{FullRes} lowers its MegaDepth AUC@$5^{\circ}$ from
$20.90$ to $18.70$. The transformer deficit is therefore, if anything,
understated, and the share of it closed by depth-warped supervision
correspondingly overstated. Comparisons \emph{among} transformer
configurations, which carry the identical recipe, are unaffected.

\subsection{Benchmarks}
\label{sec:setup:benchmarks}

\paragraph{HPatches} measures mean matching accuracy over $57$
illumination and $59$ viewpoint sequences, extracting the top $2000$
keypoints per image and reporting the proportion of mutual
nearest-neighbour matches within a pixel-error threshold. We report
threshold $1$ throughout, as the strictest and most discriminative
setting, with the illumination and viewpoint subsets separated because
the distinction carries the paper's central diagnostic.

\paragraph{MegaDepth-1500} measures relative pose accuracy over $1500$
pairs drawn from two held-out scenes. Mutual nearest-neighbour matches
feed essential-matrix estimation with RANSAC, and we report the area
under the pose-error curve at $5^{\circ}$, $10^{\circ}$ and
$20^{\circ}$. This benchmark amplifies descriptor-quality differences
relative to HPatches, since pose estimation compounds small
correspondence errors.

\paragraph{Aachen Day-Night v1.1} measures visual localisation of night
queries against a day-time reference model, extracting the top $5000$
keypoints and localising through mutual nearest-neighbour matching and
PnP in triangulation mode against the reference reconstruction. We
report the standard three thresholds, $0.25$\,m$/2^{\circ}$,
$0.5$\,m$/5^{\circ}$ and $5$\,m$/10^{\circ}$.

\begin{table*}[htbp]
\centering\footnotesize
\caption{HPatches mean matching accuracy at threshold~1, showing the
descriptor-smoothing trade-off introduced by the transformer backbone.
Restoring full spatial resolution (\textsc{FullRes}) buys $+0.072$
illumination MMA over \textsc{Base} but pays $-0.035$ on viewpoint. All
rows use top-$k=2000$ features and identical evaluation code. R2D2
(in-house) is trained by us on the same WASF data as the
transformers, without their triplet term (Section~\ref{sec:setup:baseline}),
and is our baseline; the published R2D2 and
SuperPoint rows are given for context only.}
\label{tab:vis:tradeoff}
\begin{tabular}{lccc}
\toprule
Method & Overall@1 & Illum.@1 & Viewpoint@1 \\
\midrule
\multicolumn{4}{l}{\textit{Convolutional}} \\
\quad R2D2 (in-house, our baseline) & 0.304 & 0.368 & 0.241 \\
\quad R2D2 (published, WASF\_N16)    & 0.321 & 0.395 & 0.250 \\
\quad SuperPoint                      & 0.321 & 0.393 & 0.252 \\
\midrule
\multicolumn{4}{l}{\textit{Transformer (MiT-B2 backbone)}} \\
\quad \textsc{Base}                   & 0.282 & 0.354 & 0.212 \\
\quad \textsc{FullRes}                & 0.299 & \textbf{0.426} & 0.177 \\
\bottomrule
\end{tabular}
\end{table*}

\begin{table*}[htbp]
\centering\footnotesize
\caption{Each deformable configuration against its SR-attention
counterpart, across the five deformable configurations we trained under
the two supervision pipelines of Section~\ref{sec:method:supervision} and
the three benchmarks. Entries marked ``---'' were not evaluated: of the
$15$ cells, $9$ are populated and directly paired. \textbf{The
deformable variant is worse on all $9$}, although under
depth-warped supervision its MegaDepth and Aachen deficits are within
run-to-run variation (Section~\ref{sec:results:reading}). Depth-warped
runs use the MegaDepth training scenes only. Bold marks the strongest
transformer configuration reached, which is a change of supervision rather than of architecture.}
\label{tab:vis:cross-supervision}
\begin{tabular}{llccc}
\toprule
& & HPatches & MegaDepth & Aachen \\
& Supervision & Overall@1 & AUC@$5^\circ$ & night strict \\
\midrule
\multicolumn{5}{l}{\textit{Deformable attention}} \\
\quad Variant 1 & WASF default         & 0.186 & 11.01 & --- \\
\quad Variant 2 & WASF default         & 0.203 & ---   & --- \\
\quad Variant 1 & WASF frozen-grid     & 0.227 & ---   & --- \\
\quad Variant 2 & WASF frozen-grid     & 0.217 & ---   & 46.6 \\
\quad Variant 2 & MegaDepth depth-flow & 0.237 & 24.82 & 64.4 \\
\midrule
\multicolumn{5}{l}{\textit{SR-attention}} \\
\quad \textsc{Base}    & WASF default         & 0.282 & 20.42 & 62.8 \\
\quad \textsc{FullRes} & WASF default         & 0.299 & 20.90 & 61.3 \\
\quad \textsc{Base}    & MegaDepth depth-flow & \textbf{0.283} & \textbf{25.20} & \textbf{64.9} \\
\midrule
R2D2 (in-house) & CNN baseline          & 0.304 & 32.35 & 70.2 \\
\bottomrule
\end{tabular}
\end{table*}

\begin{table*}[htbp]
\centering\footnotesize
\caption{The controlled supervision experiment. \textsc{Base} and
depth-warped SR-attention share an identical architecture, loss family and
batch size; the training data and its supervision signal differ, and
the depth-warped run received about $1.5$ times as many optimiser
iterations (${\sim}76$k versus ${\sim}50$k). Depth-warped MegaDepth
supervision, from the training scenes only, gains
$+4.78$~AUC@$5^\circ$ ($+23\%$ relative, closing $40\%$ of the
transformer-versus-CNN gap) and $+2.1$ to $+8.4$~pp on Aachen night
(about half of that gap, threshold-averaged), while leaving HPatches
essentially unchanged ($+0.001$). Under the same supervision the
deformable architecture comes within $0.38$~AUC@$5^\circ$ of
SR-attention but trails it on every metric.
\textsc{FullRes} additionally sets the stem stride to $1$ and is
included as the strongest WASF-trained configuration, not as the
architecture-matched control. Every row is the final checkpoint of its
run.}
\label{tab:vis:phase2c}
\begin{tabular}{lcccccccc}
\toprule
& \multicolumn{3}{c}{HPatches MMA} & \multicolumn{2}{c}{MegaDepth-1500 AUC} & \multicolumn{3}{c}{Aachen night} \\
\cmidrule(lr){2-4} \cmidrule(lr){5-6} \cmidrule(lr){7-9}
Method & Overall@1 & V@1 & I@1 & $5^\circ$ & $10^\circ$ & strict & med. & loose \\
\midrule
R2D2 (in-house)         & \textbf{0.304} & \textbf{0.241} & 0.368 & \textbf{32.35} & \textbf{50.44} & \textbf{70.2} & \textbf{86.4} & \textbf{95.3} \\
\midrule
\multicolumn{9}{l}{\textit{WASF supervision}} \\
\quad \textsc{Base}     & 0.282 & 0.212 & 0.354 & 20.42 & 35.75 & 62.8 & 74.3 & 88.0 \\
\quad \textsc{FullRes}  & 0.299 & 0.177 & \textbf{0.426} & 20.90 & 36.23 & 61.3 & 72.8 & 82.7 \\
\midrule
\multicolumn{9}{l}{\textit{Depth-warped MegaDepth supervision}} \\
\quad SR-attention      & 0.283 & 0.205 & 0.365 & 25.20 & 42.73 & 64.9 & 82.7 & 91.6 \\
\quad Deformable (V2)   & 0.237 & 0.137 & 0.341 & 24.82 & 41.31 & 64.4 & 78.0 & 90.6 \\
\bottomrule
\end{tabular}
\end{table*}

\begin{table*}[htbp]
\centering\footnotesize
\caption{Four interventions on the depth-warped SR-attention recipe, each holding the
supervision pipeline fixed. Only the hybrid detector moves performance
clearly: it gains on viewpoint accuracy and MegaDepth but loses on
Aachen night. The two margin changes and the larger MiT-B3 backbone
move MegaDepth by less than $2$~AUC@$5^\circ$, in both directions, and
the other benchmarks by no more than run-to-run variation. Every row is
the final checkpoint of its run.}
\label{tab:vis:interventions}
\begin{tabular}{llcccccc}
\toprule
& & \multicolumn{3}{c}{HPatches MMA} & MegaDepth & \multicolumn{2}{c}{Aachen night} \\
\cmidrule(lr){3-5} \cmidrule(lr){6-6} \cmidrule(lr){7-8}
Intervention & Axis & Overall@1 & V@1 & I@1 & AUC@$5^\circ$ & strict & loose \\
\midrule
R2D2 (in-house)            & ---          & \textbf{0.304} & \textbf{0.241} & 0.368 & \textbf{32.35} & \textbf{70.2} & \textbf{95.3} \\
Depth-warped SR-attention & reference    & 0.283 & 0.205 & 0.365 & 25.20 & 64.9 & 91.6 \\
\midrule
\quad + hybrid detector   & architecture & 0.294 & 0.230 & 0.360 & 28.42 & 63.4 & 89.5 \\
\quad + margin $0.3$      & loss recipe  & 0.291 & 0.204 & \textbf{0.382} & 26.82 & 64.9 & 92.1 \\
\quad + margin $0.1$      & loss recipe  & 0.290 & 0.218 & 0.363 & 24.01 & 66.0 & 92.7 \\
\quad + MiT-B3            & capacity     & 0.282 & 0.207 & 0.359 & 26.85 & 65.4 & 91.6 \\
\bottomrule
\end{tabular}
\end{table*}

\section{\uppercase{Results}}
\label{sec:results}

\noindent We present the results in the order in which they constrain
the question. Section~\ref{sec:results:tradeoff} establishes what the
transformer substitution actually costs, which motivates the two
candidate remedies. Section~\ref{sec:results:architecture} tests the
architectural remedy and Section~\ref{sec:results:supervision} the
supervision remedy, under conditions that isolate each one.

\subsection{The transformer backbone redistributes accuracy rather than improving it}
\label{sec:results:tradeoff}

Table~\ref{tab:vis:tradeoff} compares the two transformer
configurations against the convolutional baselines on HPatches at the
strictest matching threshold. Two observations follow.

First, the substitution is not free. \textsc{Base} reaches $0.282$
overall mean matching accuracy against $0.304$ for the convolutional
model trained by us on the same data, a deficit of $0.022$.
The two share training data, crop size and iteration budget, and differ
in the backbone and in one loss term that only the transformer carries
(Section~\ref{sec:setup:baseline}). Since that term, if anything, helps
the transformer, the deficit is attributable to the backbone rather
than to the training regime.

Second, and more usefully, the deficit is not uniform. Restoring full
spatial resolution through the stem (\textsc{FullRes}) raises
illumination accuracy to $0.426$, which exceeds \emph{every}
convolutional baseline in the table including the published R2D2 model
at $0.395$, while simultaneously depressing viewpoint accuracy to
$0.177$, the lowest figure in the table and $0.035$ below
\textsc{Base}. The transformer is therefore not globally weaker than
the convolutional baseline; it redistributes accuracy from viewpoint
change towards illumination change.

This pattern is what one expects from spatially smoothed descriptors.
Illumination pairs in HPatches are geometrically identical, so a
descriptor that pools over a wider effective support loses nothing and
gains photometric invariance. Viewpoint pairs carry a real warp, where
the same pooling blurs precisely the spatial detail that
disambiguates nearby candidates. We refer to this as the
descriptor-smoothing trade-off, and it defines the problem the rest of
the paper addresses. Two remedies are available in principle: change
the attention mechanism so that it samples adaptively rather than
uniformly, or change the supervision so that the descriptor is
trained against correspondences that penalise smoothing directly. We
test both.

\subsection{The architectural remedy does not work}
\label{sec:results:architecture}

Multi-scale deformable attention is the natural architectural
candidate. Its content-dependent sparse sampling is designed to avoid
exactly the uniform spatial pooling we identified above, and it is the
mechanism adopted by the closest published work on transformer-based
local features.

Table~\ref{tab:vis:cross-supervision} reports every deformable
configuration we trained under the two supervision pipelines against
its SR-attention counterpart under matched supervision. Five deformable
configurations across three benchmarks give $15$ cells, of which $9$
were evaluated and are directly paired. \textbf{The deformable variant
is worse on all $9$}, although two of those margins, MegaDepth and
Aachen under depth-warped supervision, are within run-to-run variation
(Section~\ref{sec:results:reading}).

The result is stable across a deliberately wide span of training
conditions: two architectural variants, a frozen-offset intervention,
and two distinct supervision pipelines. It is not an artefact of one
unlucky recipe.

One further comparison is worth stating explicitly. Under depth-warped
supervision (Table~\ref{tab:vis:phase2c}) the deformable variant trails
its SR-attention control on both HPatches subsets, but unevenly: by
$0.068$ on viewpoint pairs and by $0.024$ on illumination pairs, the
latter within run-to-run variation. The larger loss falls on the axis
that descriptor smoothing penalises, as
Section~\ref{sec:results:tradeoff} established. The mechanism introduced
to repair viewpoint accuracy does not repair it.

\subsection{The supervision remedy works}
\label{sec:results:supervision}

\textsc{Base} and the depth-warped SR-attention model share an identical
backbone, head structure, loss family and batch size. The training data
and the supervision signal derived from it are the controlled variables:
WASF image pairs are replaced by MegaDepth pairs, from the training
scenes only, whose dense correspondences are warped through measured
depth. The depth-warped run also received about $1.5$ times as many
optimiser iterations (Section~\ref{sec:setup:data}), which we cannot
separate from the supervision change.

The effect is clear on both real-image benchmarks. Relative pose AUC
at $5^{\circ}$ rises from $20.42$ to $25.20$, a gain of $4.78$ or
$23\%$ relative, which closes $40\%$ of the gap between \textsc{Base}
and the convolutional baseline. Aachen night localisation improves at
every threshold, by $2.1$, $8.4$ and $3.6$ percentage points
respectively, closing about half of the corresponding gap
when averaged across thresholds. HPatches overall accuracy is
unchanged within measurement resolution, moving from $0.282$ to
$0.283$.

Applying the same supervision change to the deformable architecture
gives a similar MegaDepth result. The deformable model under depth-warped
supervision reaches $24.82$ AUC@$5^{\circ}$, which is $0.38$ below its
SR-attention control trained on identical data, a difference within
run-to-run variation; it still trails that control on HPatches and at
every Aachen threshold. Under this supervision the choice of attention
mechanism moves MegaDepth by less than half a point, while the change of
supervision moved the SR-attention model by almost five. Supervision and
architecture are therefore not independent levers of equal weight here:
the gains are available through the supervision pipeline, and the
architectural substitution does not add to them.

\subsection{What these results do and do not establish}
\label{sec:results:reading}

Three limits should be stated plainly.

The transformer does not overtake the convolutional baseline. Our
strongest configuration on each benchmark remains below it, by $0.005$
on HPatches at threshold~$1$, $3.93$ AUC@$5^{\circ}$ below on MegaDepth-1500
and $4.2$ percentage points below at the strictest Aachen threshold. What
the supervision change buys is the closest simultaneous approach to
that baseline any transformer-backboned configuration in our study
achieved, not a new state of the art. We report it as an analysis of
where the gains in this design space are located, not as a competitive
method.

The evaluation grid is sparse. Nine of the fifteen
architecture-by-benchmark cells in
Table~\ref{tab:vis:cross-supervision} were evaluated, and several
deformable configurations were not carried to the more expensive
benchmarks once their HPatches results ruled out a viable operating
point. The $9$-of-$9$ count is over cells actually measured, and we
state it that way rather than as a proportion of the full grid.

Finally, our comparisons rest on single training runs. Two trainings
of the same deformable configuration differed by approximately one
AUC@$5^{\circ}$, by $0.017$ in HPatches overall accuracy and by
$0.035$ in illumination accuracy. The MegaDepth-1500
evaluation alone varies with a standard deviation of about
$0.6$~AUC@$5^{\circ}$ when only the order of the keypoints changes, and
each of Aachen's $191$ night queries is worth about $0.5$ percentage
points. We draw no conclusion from any difference of that size. The
supervision gain of $4.78$~AUC@$5^{\circ}$
(Table~\ref{tab:vis:phase2c}) is about five times the run-to-run
difference; the architectural gap on MegaDepth under depth-warped
supervision (Tables~\ref{tab:vis:cross-supervision}
and~\ref{tab:vis:phase2c}) is not, and we rest no claim on it.

\section{\uppercase{Why the Negative Result Holds}}
\label{sec:soundness}

\noindent A negative architectural result invites an obvious
objection: that the architecture was implemented or trained badly
rather than being unsuited to the task. We take the objection
seriously and address it with evidence from the training dynamics
themselves, which is harder to explain away than any assurance about
code quality.

\subsection{The deformable models train into their deficit}
\label{sec:soundness:trajectory}

The deformable configurations do not begin at their final level and
stay there. Variant~2 reaches $0.234$ overall matching accuracy after
five epochs, its earliest evaluated checkpoint, and then declines at
every later evaluated checkpoint to $0.203$ by epoch~$49$. Variant~1
follows the same shape: first evaluated at epoch~$25$, it falls from
$0.203$ to $0.186$.

This declining shape is informative. A model crippled by a
defective operator is poor from initialisation and stays poor; it does
not lose ground as optimisation proceeds. Part of the deficit is
something these models acquire during training, which points to the
learning dynamics of the mechanism rather than to its implementation.

\subsection{Disabling the mechanism improves the model}
\label{sec:soundness:frozen}

If the learnable sampling is implicated, freezing it should help. We
therefore retrained both variants with the sampling offsets held at
their initialisation, which removes the content-dependent sampling
that distinguishes deformable attention from a fixed-grid operator
while leaving every other component untouched.

Both frozen variants improve on all three HPatches metrics.
Variant~1 rises from $0.186$ to $0.227$ overall and Variant~2 from
$0.203$ to $0.217$. The degradation is also arrested rather than
merely offset: the frozen Variant~1 moves from $0.232$ at epoch~$25$
to $0.227$ at epoch~$49$, against $0.203$ to $0.186$ unfrozen.

Two readings follow, and we separate them carefully. As evidence about
the mechanism the result is only partial, because neither frozen
variant approaches the $0.282$ of the SR-attention baseline: the
learnable sampling is \emph{a} contributor to the deficit, not the
whole of it. As evidence about soundness, however, the result is
strong. An implementation fault in the sampling operator does not
produce a model that improves when the sampling is disabled.

\subsection{The deformable models fit better and score worse}
\label{sec:soundness:decoupling}

The sharpest evidence is the relationship between training loss and
benchmark performance. Under depth-warped supervision the deformable
configuration reaches a total training loss $14\%$ lower than its
SR-attention control in the final epoch ($0.402$ against $0.465$), while
scoring worse on every benchmark metric, by up to $33\%$ relative,
though three of those differences are within run-to-run variation.

Undertraining, an unlucky learning rate, or a misconfigured
optimiser all predict the opposite pattern: worse fit accompanying
worse benchmarks. What we observe is a model that optimises its
objective more effectively than the control and generalises to
matching less well. The deformable inductive bias is not failing to
fit the training signal; it is fitting it in a way that does not
transfer to correspondence.

\subsection{Implementation and reproduction}
\label{sec:soundness:implementation}

Our multi-scale deformable attention operator follows the Deformable
DETR reference implementation, and we verified numerical agreement
against it. The depth-warped deformable configuration reported in
Tables~\ref{tab:vis:cross-supervision} and~\ref{tab:vis:phase2c} was
trained with the verified operator: relative pose AUC at
$5^{\circ}$ of $24.82$, against $25.20$ for the SR-attention control
trained on identical data. The architectural gap has the same sign as
every other informative pair in
Table~\ref{tab:vis:cross-supervision}; on MegaDepth its size is within
run-to-run variation.

\subsection{Breadth of the evidence}
\label{sec:soundness:breadth}

Two further properties of the study bear on the objection. First, the
comparison spans five configurations and two supervision pipelines
rather than one recipe, as
Table~\ref{tab:vis:cross-supervision} sets out; a training failure
specific to one configuration would not survive that span. Second, the
convolutional baseline is our own retraining on the same data and a
comparable iteration budget rather than a published number, so
the comparison cannot be attributed to a difference in training
data between our models and someone else's; the one difference in
objective is set out in Section~\ref{sec:setup:baseline}.

\subsection{What limits the supervision result}
\label{sec:soundness:ceiling}

The supervision result of Section~\ref{sec:results:supervision} does
not reach the convolutional baseline either, and
Table~\ref{tab:vis:interventions} reports what we tried in order to
close the remainder. Four interventions were applied to the depth-warped SR-attention
recipe with the supervision pipeline held fixed, along the
architecture, loss-recipe and capacity axes.

Only the hybrid detector moves performance clearly. Substituting a
hybrid convolutional detector trunk raises viewpoint accuracy by
$0.025$ and MegaDepth pose accuracy by $3.2$~AUC@$5^{\circ}$, but loses
$1.5$ and $2.1$ percentage points at the strictest and loosest Aachen
thresholds. The two triplet-margin changes and the move from MiT-B2 to
MiT-B3 change MegaDepth by less than $2$~AUC@$5^{\circ}$, in both
directions, and the other benchmarks by no more than run-to-run
variation. Increasing backbone capacity therefore does not clearly
help, which is not what a capacity-limited ceiling predicts.

Taken together, none of these interventions closes the remaining gap,
and we do not identify what limits the supervision result. They are
consistent with the paper's main finding nonetheless: in this design
space the supervision pipeline is the one change we tested that
improves MegaDepth and Aachen together. The hybrid detector improves
MegaDepth at a cost on Aachen, and the capacity and margin changes do
not clearly move performance.

\section{\uppercase{Conclusions}}
\label{sec:conclusions}

\noindent We set out to separate two variables that the strongest
published result in transformer-based local features changes together.
Holding a single pipeline fixed and varying one axis at a time, we find
that the two are not of comparable weight. Substituting multi-scale
deformable attention for spatially-reduced attention was
harmful on all nine paired measurements spanning two supervision
pipelines, while substituting depth-warped correspondence supervision on
an unchanged architecture gained about $4.8$~AUC@$5^{\circ}$ and closed
about $40\%$ of the gap to a convolutional baseline we retrained on the
same data. In this design space, effort spent on the supervision
pipeline returns considerably more than effort spent on the attention
mechanism.

The negative result is supported by evidence that argues against
an implementation or training failure: the deformable models lose
accuracy as training proceeds, Variant~1 improves clearly
when its defining content-dependent sampling is disabled, and under
depth-warped supervision the deformable model reaches a lower training
loss than its control while scoring worse on every benchmark metric. A model that fits
its objective better and generalises to matching worse is not an
undertrained model.

Three limitations bound these conclusions. Our comparisons rest on
single training runs, and we measured a run-to-run difference of
approximately one AUC@$5^{\circ}$ between two trainings of the same
configuration, so we draw no conclusion from any narrower ordering;
the architectural gap on MegaDepth under depth-warped supervision is
one such ordering. The
evaluation grid is sparse, nine of fifteen cells, because
configurations ruled out on the cheapest benchmark were not carried to
the more expensive ones. And the supervision result does not reach the
convolutional baseline, its size is approximate because the
depth-warped runs trained for more iterations and the baseline lacks
the triplet term, and none of the interventions we tried closes the
remaining gap.

Finally, the scope of the negative result deserves care, and our own
central finding is the reason. If supervision dominates architecture,
then an architectural verdict reached under one family of supervision is
a supervision-scoped verdict. We tested two supervision pipelines, which is what gives
the result its force, but it remains a claim about deformable attention
under R2D2-style objectives in this pipeline, not a claim about
deformable attention in general. The natural next step is the converse
experiment: holding the supervision at the correspondence-supervised
recipes under which deformable attention was originally reported to
succeed, and varying the architecture within them.

\section*{\uppercase{Acknowledgements}}

\noindent 
This publication has emanated from research supported in part by grants from Taighde Éireann - Research Ireland under Grant numbers 18/CRT/6049 and 16/RI/3399. The author used Claude
(Anthropic)~\cite{anthropic_claude} as coding and writing assistant. The research design, the experiments, the results and
their interpretation are the author's own.
\vfill
\bibliographystyle{apalike}
{\small
\bibliography{references}}

\end{document}